%% file: main.tex
\documentclass[10pt,twocolumn,letterpaper]{article}
\usepackage{cvpr}              % To produce the CAMERA-READY version
\usepackage{graphicx}
\usepackage{amsmath}
\usepackage{amssymb}
\usepackage{booktabs}
\usepackage{soul}
\usepackage{makecell}

\usepackage{hyperref}
\usepackage{url}
\usepackage{booktabs}       % professional-quality tables
\usepackage{amsfonts}       % blackboard math symbols
\usepackage{nicefrac}       % compact symbols for 1/2, etc.
\usepackage{microtype}      % microtypography
\usepackage{xcolor, colortbl}         % colors
\usepackage{fancynum}

\usepackage{xspace}
\usepackage{amsthm, amsmath, amssymb}
\usepackage{cmap}
\usepackage{caption}
\usepackage{pifont}
\usepackage{subcaption}
\usepackage{bibentry}
\usepackage{wrapfig}
\usepackage{mathrsfs}
\usepackage{multicol}
\usepackage{multirow}
\usepackage{threeparttable}

\usepackage{nicefrac}       % compact symbols for 1/2, etc.
\usepackage{microtype}      % microtypography

\newcommand{\vus}{\textit{vs}. }
\usepackage[capitalize]{cleveref}
\crefname{section}{Sec.}{Secs.}
\Crefname{section}{Section}{Sections}
\Crefname{table}{Table}{Tables}
\crefname{table}{Tab.}{Tabs.}

\def\cvprPaperID{1986} % *** Enter the CVPR Paper ID here
\def\confName{CVPR}
\def\confYear{2024}

\begin{document}

\title{Can I Trust Your Answer? Visually Grounded Video Question Answering}

\author{Junbin Xiao \quad Angela Yao\thanks{corresponding author} \quad Yicong Li \quad Tat-Seng Chua \\
Department of Computer Science, National University of Singapore \\ 
{\tt\small{\{junbin,ayao,chuats\}@comp.nus.edu.sg}, liyicong@u.nus.edu}
}

\twocolumn[{
\renewcommand\twocolumn[1][]{#1}
\maketitle
\vspace*{-5.5mm}
\centering
\iftoggle{cvprfinal}{\includegraphics[width=1.0\linewidth, height=0.15\linewidth]{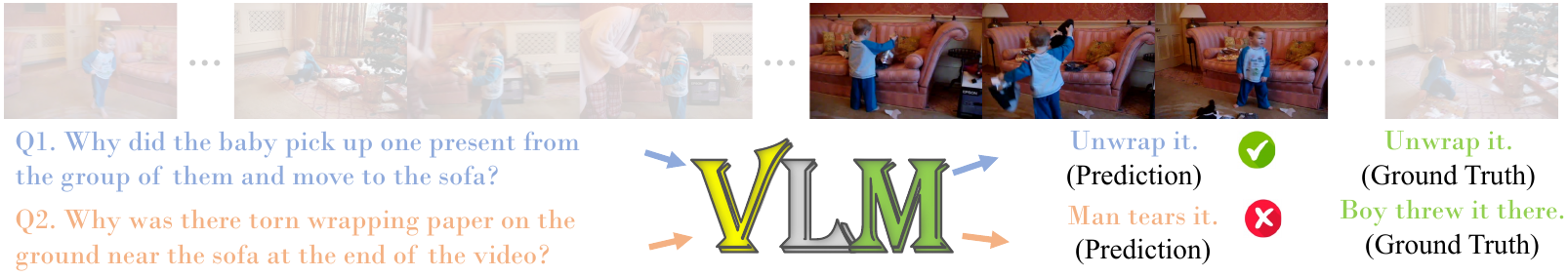}}{\includegraphics[width=1.0\linewidth, height=0.15\linewidth]{figures/intro-exp.pdf}} \\
\vspace{1em}
\iftoggle{cvprfinal}{\includegraphics[width=1.0\linewidth, height=0.2\linewidth]{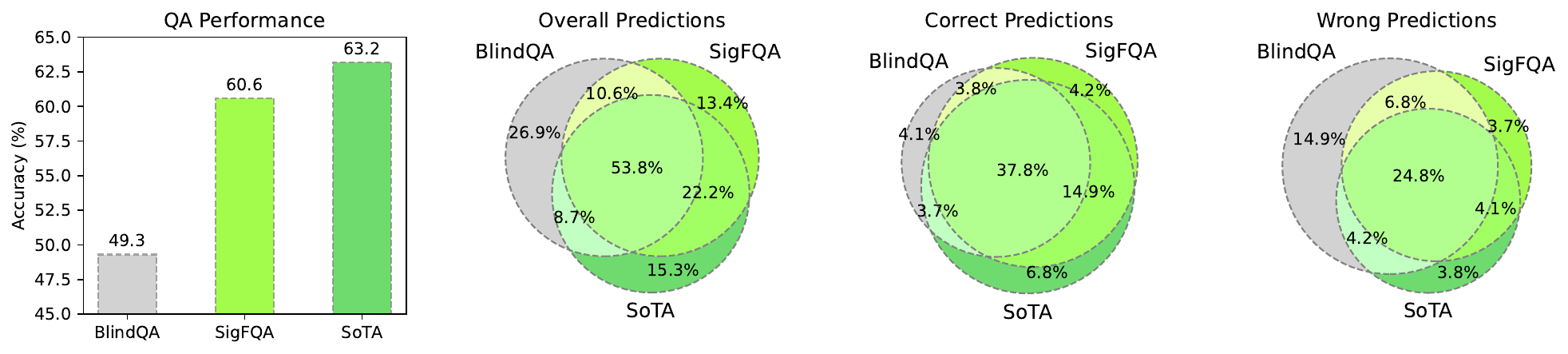}}{\includegraphics[width=1.0\linewidth, height=0.2\linewidth]{figures/intro-res.pdf}}
\vspace{-0.5cm}
\captionof{figure}{~\textbf{Top}: Real predictions of VQA models (BlindQA, SigFQA and SoTA) on NExT-QA \cite{xiao2021next}. All the models correctly answer \texttt{Q1} but wrongly answer \texttt{Q2}, albeit the two questions sharing visual evidence (\emph{the boy unwraps the present and throws the wrapping paper}). \textbf{Bottom}: Overlap in model predictions. BlindQA: A pure language model 
(\ie~RoBERTa \cite{liu2019roberta}) 
fine-tuned with question-answer text.
SigFQA: An image-text model 
(\ie~CLIP \cite{radford2021learning}) 
using only the center video frame. % as input.
SoTA: Temp[CLIP] (Sec.~\ref{sec:exp}) model using 32 video frames. 
The analyses indicate that the models may not learn from causal visual content but more likely from language short-cut and irrelevant visual context.}
\label{fig:intro-vil}
\vspace*{5mm}
}]

\maketitle

%%%%%%%%% ABSTRACT
\begin{abstract}

We study visually grounded VideoQA in response to the emerging trends of utilizing pretraining techniques for video-language understanding. Specifically, by forcing vision-language models (VLMs) to answer questions and simultaneously provide visual evidence, we seek to ascertain the extent to which the predictions of such techniques are genuinely anchored in relevant video content, versus spurious correlations from language or irrelevant visual context. Towards this, we construct NExT-GQA -- an extension of NExT-QA with 10.5$K$ temporal grounding (or location) labels tied to the original QA pairs. With NExT-GQA, we scrutinize a series of state-of-the-art VLMs. Through post-hoc attention analysis, we find that these models are extremely weak in substantiating the answers despite their strong QA performance. This exposes the limitation of current VLMs in making reliable predictions. As a remedy, we further explore and propose a grounded-QA method via Gaussian mask optimization and cross-modal learning. Experiments with different backbones demonstrate that this grounding mechanism improves both grounding and QA. With these efforts, we aim to push towards trustworthy VLMs in VQA systems. Our dataset and code are available at \url{https://github.com/doc-doc/NExT-GQA}. 
\end{abstract}

\input{sections/intro}

\input{sections/related}

\input{sections/benchmark}
\input{sections/method}
\input{sections/experiment}
\input{sections/conclusion}

%%%%%%%%% REFERENCES
{\small
\bibliographystyle{ieee_fullname}
\bibliography{main}
}

\clearpage
\appendix
\input{sections/appendix}

\end{document}

%% file: sections/intro.tex
\section{Introduction}\label{sec:intro}
Video Question Answering (VideoQA) has recently emerged as a golden testbed to develop vision-language models (VLMs), especially foundation VLMs pretrained at scale on multi-modal web corpora \cite{li2020hero,lei2021less,yang2021just,zellers2021merlot,alayrac2022flamingo,ye2022hitea,wang2022internvideo,fu2022empirical}. 
Despite significant advancements
in QA performance, a fundamental concern arises -- whether or to what extent are
the answers of such techniques %are truly 
grounded on the relevant visual content? Alternatively, are they relying on the \emph{language short-cut} for the use of powerful language models \cite{piergiovanni2022video,yang2022zero,zhao2023learning, wang2023paxion,yu2023self,ko2023large,zhang2023video,zhang2023llama,lin2023video} or \emph{spurious vision-language correlation} captured via cross-modal pretraining \cite{raffel2020exploring,xu2021videoclip}?

For example, Fig.~\ref{fig:intro-vil}(Top) %we find that 
shows that existing VLMs are inclined to answer questions with \emph{language-biased} predictions, \eg, ``\texttt{unwrap (Q1: present)}'' and ``\texttt{tear (Q2: paper)}''. Fig.~\ref{fig:intro-vil}(Bottom) shows that the overall predictions of
SoTA VLMs overlap the predictions of standalone language models (BlindQA), \ie, models without visual inputs, by 62.5\%. %(53.8+8.7).
In fact, the BlindQA counterpart shares 66\% 
% ((37.8+3.7)/63.2) 
of correct predictions
and also 79\% of the wrong predictions in SoTA VLMs.
% ((24.8+4.2)/36.9) 
% of the wrong predictions. 
The overlap %gets much heavier 
increases by injecting a coarse visual signal from a single frame \cite{buch2022revisiting,lei2022revealing} into the language model; as our later analysis will show, %s that, 
%most of the time, 
this frame often lies outside the key moments of the correct answers.

Given these findings, a natural question arises -- to what extent 
are the predictions of current VLMs
grounded on the video content, and more precisely on the \emph{relevant} parts?
To answer this, we propose to study \textbf{visually grounded VideoQA}.  Grounded VQA requires VLMs to answer the questions and simultaneously output the relevant video moments to support the answers. 
Earlier works have explored grounded QA under full supervision \cite{lei2018tvqa,lei2020tvqa+}, but we target %the analysis of 
visual explanability in VideoQA and thus define the task under weak-supervision, which is the first of its kind. 

To accomplish the goal, we construct the NExT-\underline{G}QA (short for Grounded) dataset by extending the NExT-QA dataset \cite{xiao2021next} with 10.5$K$ temporal labels of start and end timestamps for the QA pairs in the validation and test sets. The labels are manually annotated and checked to be key for comprehending the questions and determining the correct answers. With NExT-GQA, we examine a series of recent high-performing VLMs, including
task-specific architectures without pretraining~\cite{xiao2022vgt} and  
pretrained models with either image-text or video-text data \cite{radford2021learning,fu2022empirical} and those using frozen large language models (LLMs) \cite{yang2022zero,yu2023self}. Our findings reveal that all these models struggle to predict visually grounded answers, despite their strong QA performance. 
For example, the SoTA model \cite{yang2022zero} achieves QA accuracy of 69\%, but only 16\% of the correctly predicted answers are grounded in the video. In contrast, humans can ground 82\% out of the 93\% of the correctly answered questions.
Such clear discrepancy underscores the need for continued research efforts.

As a pioneering solution, we propose a temporal grounding approach which can be easily applied to existing VLMs for
visually grounded VideoQA. 
Specifically, our approach learns differentiable Gaussian masks along the temporal dimension of the videos, 
by optimizing light-weight transformer layers under both VQA and question-video (QV) supervisions, without the need for temporal labels. 
Experiments with different QA backbones demonstrate that our approach effectively improves
video grounding and question answering as well.
The improvement is especially significant on a subset of questions that necessitate video understanding and temporal grounding.

To summarize our contributions: 1) we conduct the first study of weakly grounded VideoQA, and release the NExT-GQA benchmark, to facilitate research on more trustworthy VLMs; 2) we comprehensively analyze a wide range of advanced VLMs and reveal their
limitation in performing visually grounded QA; 3) we propose
a simple yet effective grounding mechanism which not only enhances existing VLMs in visual grounding but also contributes to new SoTA QA performance, \eg, 73.1\% on NExT-QA test set.

%% file: sections/related.tex
\section{Related Work}\label{sec:relatedwork}

\paragraph{Benchmarks} Grounded VQA with full supervision has been studied in both image \cite{zhu2016visual7w,chen2022grounding} and video \cite{lei2018tvqa,lei2020tvqa+} domains. Recently, weakly-supervised grounding has received increasing attention in ImageQA \cite{khan2021found,khan2022weakly} and 
video grounding \cite{gao2017tall,mithun2019weakly}. 
Nonetheless, to our best knowledge, there is no work for weakly-grounded VideoQA. Also, existing supervised benchmarks are either biased towards localizing subtitles in TV shows (\eg~TVQA \cite{lei2018tvqa}) or limited to few objects (\eg~VidSTG \cite{zhang2020does}). Thus, they are not ideal benchmarks for visual evidence grounding.
\vspace{-0.4cm}
\paragraph{Techniques} Strong \textbf{VideoQA} methods %performance is dominated by 
are predominantly banked on transformer \cite{vaswani2017attention} and pre-training %techniques
~\cite{radford2021learning}. 
The popular transformer architectures follow either shared \cite{lei2021less,wang2022all}, dual \cite{yang2021just,xu2021videoclip,xiao2022vgt,xiao2023contrastive} or stacked \cite{li2020hero,fu2022empirical,yang2022zero} implementations, and pre-training is done with image-text~\cite{lei2021less}, video-text \cite{yang2021just,li2020hero,zellers2021merlot,zellers2022merlot} or both~\cite{fu2022empirical} forms of data. Notably, all these VLMs use powerful language models (\eg, BERT \cite{devlin2018bert}, T5 \cite{raffel2020exploring}, GPT \cite{brown2020language}, LLaMA \cite{touvron2023llama} or their successors) for text encoding and focus on improving QA while ignoring visual evidence grounding. Some recent works~\cite{li2022invariant,li2022equivariant,li2023trans,li2023discovering,chen2021explainable,qian2022locate,yu2023self} have begun to ground key frames or objects for VideoQA. Yet, they still aim to improve QA accuracy, and thus the grounded contents may not be the \emph{actual} evidences since they do not evaluate grounding.
For weakly-supervised \textbf{video grounding}, typical approaches extract temporal proposals and rank the proposals according to their similarities with the language query \cite{gao2019wslln,mithun2019weakly,zhang2020counterfactual,lin2020weakly}. 
Despite their effectiveness, these two-stage approaches are notorious for 
inefficient and sub-optimal for multi-granular temporal modelling. More recent research \cite{zheng2022cnm,zheng2022weakly} points to the superiority of end-to-end Gaussian mask learning. In light of this, we design a simple yet effective Gaussian mask learning module for grounding in VideoQA. Unlike previous works~\cite{zheng2022cnm,zheng2022weakly} that design small transformer models and hand-craft negative visual proposals for contrastive learning, we integrate Gaussian mask learning into large VLMs and optimize its parameters via question-answering and video-question grounding. 
\vspace{-0.4cm}
\paragraph{Language Priors} Our work is also related to efforts in preventing language priors and other spurious correlations. Goyal \etal \cite{goyal2017making} construct VQAv2 to prevent language priors in VQA by pairing the questions with additional images that carry similar contents but with different answers. Niu \etal \cite{niu2021counterfactual} and Guo \etal \cite{guo2021loss} alleviate language priors by regularizing the prediction scores. Zheng \etal \cite{zeng2023x} develop X2-VLM via multi-grained vision-language pretraining for better spatial grounding. These works discourage short-cut learning in the image side by either collecting new data, designing tailored-made learning methods, or focusing on spatial grounding. Our primary contribution lies in defining the weakly-grounded VideoQA task to encourage more interpretable and trustworthy techniques.

%% file: sections/benchmark.tex
\section{NExT-GQA Dataset}\label{sec:benchmark}
% Recall that our goal is to provide a benchmark for visual evidence grounding in VideoQA. In this section, we introduce how to construct the grounded VideoQA dataset, \ie, NExT-GQA.
\subsection{Dataset Construction and Analysis}

\textbf{Data Source}. 
We choose NExT-QA \cite{xiao2021next} as our data source to augment with
temporal labels. Most of the other VideoQA datasets~\cite{jang2017tgif,xu2017video} are not suitable because they feature short videos (3 $\sim$ 15s) already trimmed around the relevant content. %and do not require temporal grounding to answer the questions. 
NExT-QA %mainly defines 
has three different types of questions: Causal (``\texttt{why/how}''), Temporal (``\texttt{before/when/after}'') and Descriptive (``\texttt{what/who/where}''). We exclude the descriptive questions %as we find that they mostly ask the 
because they mostly pertain to global video 
content (\eg, ``\texttt{what event?}'') or their 
answers can be found almost throughout the whole video (\eg, ``\texttt{where is?}''). In addition, we only %annotate temporal 
label %for the QAs in 
the validation and test sets since we aim for a weakly-supervised setup. % grounding. 
As a result,\fnum{11378}QA pairs drawn from\fnum{1570}videos \cite{shang2019annotating} are to be annotated.

\textbf{Label Collection}.
We invite undergraduate students 
for annotation (using Elan \cite{elan65}) and train them with our demo annotations together with some trial examples following specific criteria (see Appendix~\ref{sec:appdset}) 
before the actual annotation exercise. To guarantee quality and reduce subjectiveness, each QA pair is annotated by at least two people. The final temporal label is determined by an additional check and refinement %on the basis 
of the two accepted annotations. 
% The annotation is facilitated by Elan \cite{elan65}. 
The entire exercise lasted around 2 months with a team of 30 annotators. Eventually, we collect \fnum{10531} valid temporal segments corresponding to \fnum{8911} QA pairs and \fnum{1557} videos. Detailed statistics are presented in Tab.~\ref{tab:nextgqa}.
\setlength{\tabcolsep}{4pt}
\begin{table}[t!]
  \caption{Statistics of NExT-GQA.}
  \vspace{-0.3cm}
  \label{tab:nextgqa}
  \centering
  \small
  \scalebox{0.8}{
  \begin{tabular}{lcccccc}
    \toprule
    Split & \#Vid. & \#Que. & \#Seg. & Seg. Dur.(s) & Vid. Dur.(s) & Ratio (S.$/$V.) \\
    \midrule
    Train & \fnum{3860} & \fnum{34132} & - & - & 44.9 & - \\
    Val & 567  & \fnum{3358}   & \fnum{3931} & 7.3 & 42.2 & 0.2 \\
    Test     & 990 & \fnum{5553}    & \fnum{6600} & 6.7 & 39.5 & 0.2  \\
    \bottomrule
  \end{tabular}
  }
  \vspace{-0.4cm}
\end{table}

\begin{figure}[t!]
     \centering
     \begin{subfigure}{0.49\textwidth}
         \centering
         \includegraphics[width=\textwidth]{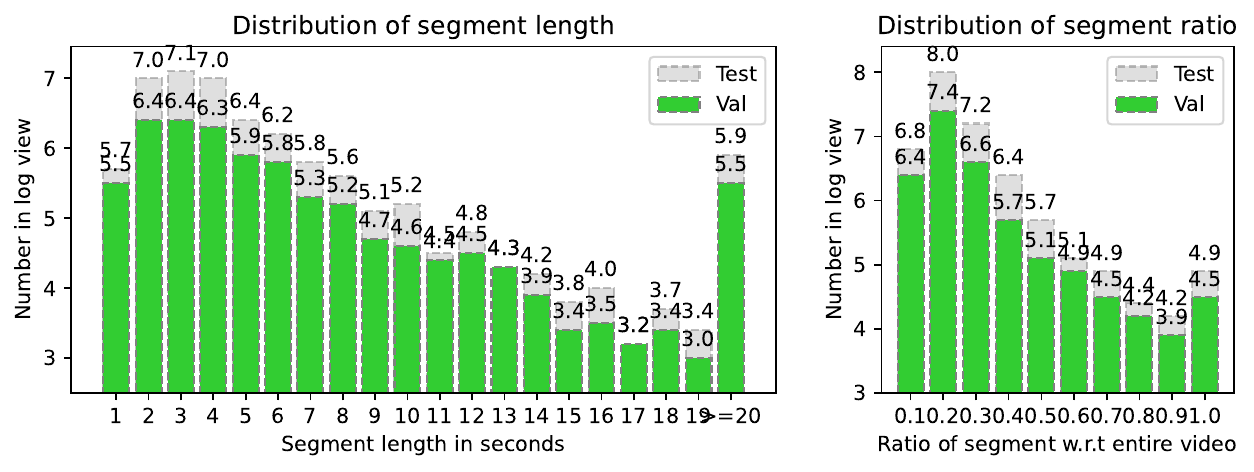}
         \vspace{-0.6cm}
         \caption{}
         \label{fig:seg-bar}
     \end{subfigure}
     \hfill
     \begin{subfigure}{0.49\textwidth}
         \centering
         \includegraphics[width=\textwidth]{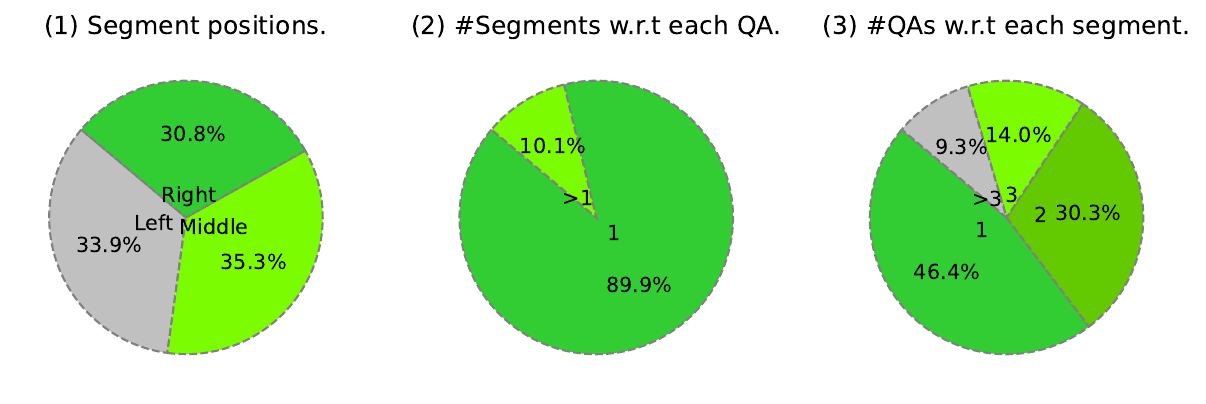}
         \vspace{-0.6cm}
         \caption{}
         \label{fig:seg-pie}
     \end{subfigure}
     \vspace{-0.5cm}
     \caption{Distribution of temporal segments.}
     \label{fig:seg-dis}
     \vspace{-0.4cm}
\end{figure}

\textbf{Label Analysis}.
Fig.~\ref{fig:seg-bar}(left) shows that most of the segments last for less than 15s, and with an average duration of 7s (Tab.~\ref{tab:nextgqa}) which is short compared to the video length ($\sim$40s). In fact, the ratio  
reflected in Fig.~\ref{fig:seg-bar}(right) shows that most of the segments occupy less than half (0.5) length of the videos, and the average ratio is merely 0.2 (Tab.~\ref{tab:nextgqa}. This ratio is slightly low, compared to that of 0.3 for both ActivityNet-Caption \cite{krishna2017dense} and Charades-STA \cite{gao2017tall}. Moreover, Fig.~\ref{fig:seg-pie}(1) shows that the segments are evenly distributed in the left, middle and right parts of the video. Fig.~\ref{fig:seg-pie}(2) shows that near 90\% of the QAs ground on a single temporal segment. Conversely, Fig.~\ref{fig:seg-pie}(3) shows that each segment often corresponds to 1 or 2 QAs (Here two segments are considered the same if their IoU $>$ 0.5). To better understand the dataset, we show two examples in Appendix Fig.~\ref{fig:seg-exp}.
% and more are found in the Appendix.

\subsection{Comparison with Existing Benchmarks}
We highlight the uniqueness of NExT-GQA by comparing it with other relevant benchmarks in Tab.~\ref{tab:benchmark}. 

\textbf{NExT-GQA \vus NExT-QA}. %Compared with 
NExT-QA \cite{xiao2021next} %that aims at 
targets the prediction of text answers. 
NExT-GQA differs in two major aspects: 1) it 
%goes beyond that to 
provides visual evidence to support the answers, and 2) it extends the VQA setup by allowing visual answers. This satisfies more real-world applications and additionally helps to better diagnose model performance. For example, is a prediction wrong because the model failed to localize the relevant video contents, or because it could not convert the localized video contents into a text answer? 
NExT-GQA is also more challenging, because: 1) the models %not merely 
need to achieve multiple goals (\ie, grounding and QA) %simultaneously but also should 
and \emph{maintain their consistency}, and 2) the set of questions are relatively harder to answer by focusing on local video moments in untrimmed long videos. This also differs from major VideoQA benchmarks that focus on trimmed (short) video understanding \cite{xu2017video,jang2017tgif,wu2021star}.

\textbf{NExT-GQA \vus Video Grounding (VG) Benchmarks}. %Compared with 
Video grounding~\cite{gao2017tall,krishna2017dense} %which 
aims to find a video moment described by a declarative sentence. NExT-GQA
shares core challenges, \ie, \emph{cross-modal correspondence learning} and \emph{multi-granular temporal modelling}, while featuring some unique aspects.
\textbf{First}, the questions feature visual content for grounding which is not explicitly stated in the text, such as ``\texttt{a baby falls and cries.}'' \vus ``\texttt{why did the baby cry?}''. To answer the questions, the models not only need to find the described video moments (\eg, ``\texttt{baby cries}'') but also should be capable of refining the moment to enclose the answer (\eg, ``\texttt{baby falls}''). This may ask for \emph{temporal and causal relationship} reasoning. \textbf{Second}, the video backgrounds are relatively monotonous with little scene change. Accordingly, the temporal segments of QA pairs are often more fine-grained than those in VG benchmarks. 
\textbf{Notably}, NExT-GQA prioritizes finding visual evidence to support the answers. This means that any individual frame or moment that sufficiently tells the answer should be considered as a valid grounding instead of retrieving all of the video contents that match the query. 
This is reflected in our selection of intersection over prediction (\textbf{IoP}) as an evaluation criterion. That is, a correct grounding depends on whether the predicted segment falls into the labelled segment but is not necessarily an exact match.
\setlength{\tabcolsep}{6.2pt}
\begin{table}[t!]
  \caption{Benchmark comparison. GD: Grounding. MM: Multimodal. Acc: Accuracy. IoU/P: Intersection over Union/Prediction.}
  \label{tab:benchmark}
  \vspace{-0.3cm}
  \centering
  \small
  \scalebox{0.8}{
  \begin{tabular}{lccccc}
    \toprule
    Datasets & GD & QA & Weak Sup. & Goal & Eval \\
    \midrule
    ActNet-Cap \cite{krishna2017dense} & $\surd$ & $\times$ & $\surd$ & VG & IoU   \\
    Cha-STA \cite{gao2017tall} & $\surd$  & $\times$  & $\surd$ & VG & IoU \\
    TVQA \cite{lei2018tvqa}    & $\surd$ & $\surd$    & $\times$ & MMVQA & Acc, IoU  \\
    VidSTG \cite{zhang2020does}    & $\surd$ & $\surd$    & $\times$ & VG & IoU \\
    NExT-QA \cite{xiao2021next} & $\times$ & $\surd$    & $\times$ & VQA & Acc \\
    \midrule
    NExT-GQA   & $\surd$ & $\surd$   & $\surd$ & Trust VQA & Acc, IoP, IoU \\
    \bottomrule
  \end{tabular}
  }
  \vspace{-0.4cm}
\end{table}

\textbf{NExT-GQA \vus Supervised benchmarks}. Fully-supervised benchmarks \cite{lei2018tvqa, lei2020tvqa+} provide temporal annotations for training data; the labels can resolve reference ambiguities in the questions or improve QA performance with well-localized visual inputs. 
NExT-GQA differs from them by seeking to identify visual evidence that explains the answers with QA supervision alone. It is worth mentioning that directly applying the fully-supervised benchmarks for weakly grounding does not suit our goal, because these benchmarks are either biased to text localization \cite{lei2018tvqa} or the answers are a limited set of, \eg~80 objects \cite{zhang2020does}. Additionally, we focus on weakly-supervised \emph{temporal} grounding and leave \emph{spatio-temporal} grounding for future exploration. Our consideration is that fine-grained \emph{spatio-temporal} grounding \cite{zhang2020does} is currently more challenging than question-answering, especially in the weak supervision setting \cite{xiao2020visual}, and would derail the main goal of VQA.

%% file: sections/method.tex
\section{Weakly-Supervised Grounding in VideoQA}\label{sec:method}
% In this section, we describe our solution towards weakly-grounded VideoQA. At the core of our method is a question grounding module that can be seamlessly integrated into existing VideoQA models to achieve grounding as well as improved question-answering.

\begin{figure*}[t!]
  \centering
  \includegraphics[width=1.0\linewidth]{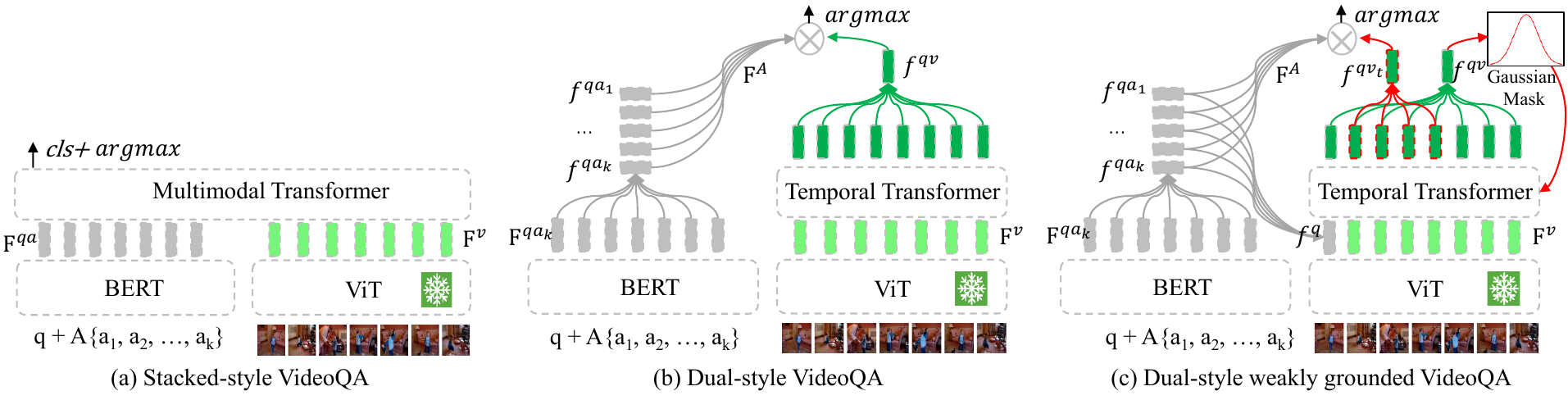}
  \vspace{-0.5cm}
  \caption{Illustration of stacked (a) and dual (b) style Transformer architecture for VideoQA. (c) Our example of dual-style weakly-grounded VideoQA. Note that the grounding part is identical for stacked-style implementation. }
  % The position and modality embeddings, along with the special tokens, are omitted for brevity. }
  \label{fig:solution}
  \vspace{-0.3cm}
\end{figure*}

\textbf{VideoQA}.
%To better convey our solution, 
We %briefly recap the
first give an overview of the typical approaches to VideoQA, focusing on %the 
transformer-based methods due to their % for their dominated 
superior performance. Given a video $v$ 
% represented by $F^v=\{f_i^v\}_{i=1}^n$ 
and a question $q$ 
% represented by $F^q=\{f_j^q\}_{j=1}^m$ ($n$ and $m$ denote the lengths of the video and the question respectively.)
, the goal of VideoQA is to predict a correct answer 
$a^*$
from a set of candidate answers $A$. Depending on the task setting, 
$A$ can be given by multiple choices accompanying each question \cite{xiao2021next,xiao2022video} (multi-choice), or by a global answer set \cite{xu2017video} to all questions (open-ended). Note that SoTA transformer-%style solutions 
methods~\cite{yang2021just,xiao2022vgt,yang2022zero,fu2022empirical} formulate and solve both multi-choice QA and open-ended QA in a unified formulation:
\begin{equation}\label{eqn:vqa}
    a^*=\text{argmax}_{a\in A}~\Psi (a | v, q, A),
\end{equation}
in which the mapping $\Psi$ is typically %popularly 
realized as either shared \cite{lei2021less,wang2022all}, stacked~\cite{fu2022empirical,li2020hero,yang2022zero} or dual~\cite{radford2021learning,yang2021just,xiao2022vgt} transformer. In this work, we primarily study the behaviour of the stacked- (Fig.~\ref{fig:solution}a) and dual-transformer (Fig.~\ref{fig:solution}b)  architectures for their relatively better performance. 

\textbf{Weakly Grounded VideoQA}.
Aside from answering questions, weakly-grounded VideoQA requires the models to explicitly estimate a QA-relevant video segment to serve as visual evidence. We introduce below three model-agnostic solutions to achieve this goal:

\textbf{Post-hoc (PH)}. Intuitively, relevant temporal segments can be found through a \emph{post-hoc} analysis of the temporal attention, \ie, identifying the segment or frame with the maximal attention value and then thresholding around it to obtain a time interval. To that end, we use \emph{attention-pooling} to summarize the outputs from the temporal transformers for dual architectures. 
% We have also tried to \emph{prepend} a token to summarize the token information in dual-style transformers but found that this practice converges much slower than \emph{attention-pooling} and the final performances are similar. 
For stacked architectures, we directly return the averaged multi-head attention values corresponding to the prediction token. 
\begin{figure*}[t!]
     \centering
     \begin{subfigure}{0.47\textwidth}
         \centering
         \includegraphics[width=\textwidth]{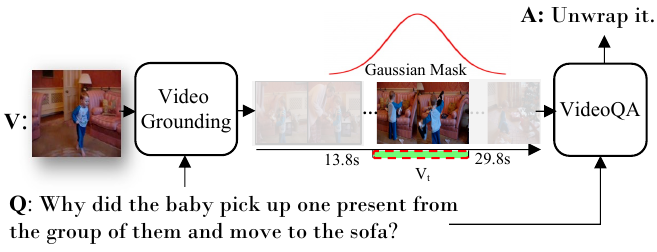}
         \caption{Framework of weakly-grounded VideoQA.}
         \label{fig:framework}
     \end{subfigure}
     \hfill
     \vspace{0.5em}
     \begin{subfigure}{0.47\textwidth}
         \centering
         \includegraphics[width=\textwidth]{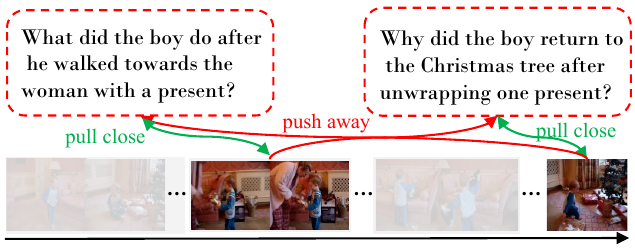}
        \caption{Cross-modal self-learning of GQA.}
         \label{fig:cst}
     \end{subfigure}
      \vspace{-0.3cm}
     \caption{Illustration of the framework (a) and our NG+ solution (b) for weakly-grounded VideoQA.}
     \label{fig:main}
     \vspace{-0.5cm}
\end{figure*}

\textbf{Naive Gaussian (NG)}. The \emph{post-hoc} approach is designed to 
analyze the models, but not influence their predictions. More favourably, we propose to explicitly incorporate a video grounding mechanism into VideoQA. We illustrate the framework in Fig.~\ref{fig:framework}, and reformulate Eqn.~\ref{eqn:vqa} as
\begin{equation}\label{eqn:gvqa}
a^*,~t^* = \text{argmax}_{a\in A}~\Psi (a | v_t, q, A)\Phi (t | v, q),
\end{equation}
in which the grounding module $\Phi$ firstly estimates the key moment specified by $t$ and thereafter the QA module $\Psi$ takes the more localized video content $v_t$ for answer prediction. To enable end-to-end learning, $t$ is represented by differentiable Gaussian weights over the entire video sequence, \ie, $t\sim N(\mu, \sigma^2)$, where $\mu$, $\sigma\in[0, 1]$ are two learnable Gaussian parameters corresponding to the mean and standard deviation. During inference, the grounding can be achieved by the confidence interval $t=(\mu-\gamma\sigma, \mu+\gamma\sigma)*d$, where $\gamma$ is a hyper-parameter to control the width of the confidence interval and $d$ denotes the duration of the video. 

Fig.~\ref{fig:solution}c shows a dual transformer instantiation of this naive % weakly-grounded VideoQA 
solution. The difference with respect to the original VideoQA counterpart (Fig.~\ref{fig:solution}b) lies in a Gaussian mask prediction head, along with a Gaussian weighted token learning and aggregation stage (details in Appendix \ref{sec:appimp}).
We find that this approach effectively learns and outputs grounding information. Nevertheless, the improvements over a \emph{post-hoc} solution are limited due to the weak QA supervision.

\textbf{NG+}. In light of the naive Gaussian results, we further design an auxiliary objective with cross-modal self-supervision to regularize the VQA objective towards more visually grounded QA. Specifically, for each question $q^+$, we treat the corresponding grounding hypothesis $v_t$ as an anchor point and pull it close to $q^+$ while pushing it away 
from other questions $Q^-$ in the feature space. The negative set $Q^-$ includes: 1) other questions defined in the same video as \emph{hard} negatives, since a large portion (near half) of questions each invokes unique video moment for answer (Fig.~\ref{fig:seg-pie}(3)); 2) questions sampled from other videos to ensure the sufficiency and diversity of negative samples. 
Moreover, we enrich 10\% of the positive questions by rephrasing each question (using GPT-4 \cite{OpenAI2023GPT4TR}) with a maximum of 5 additional questions to form $Q^+$. 
It is worth noting that there is 
only one positive question at each training iteration and the enriched positive questions are randomly picked to substitute the original one for data augmentation. Such form of contrast 
is thus implemented as classification by also  
fixing the number of negative questions to be identical to that of distractor answers. Thereby, 
our final solution is:
\begin{equation}\label{eqn:cgvqa}
\begin{aligned}
a^*,~t^* = \underbrace{\text{argmax}_{a\in A}~\Psi (a | v_t, q^+, A)\Phi (t | v, q^+)}_{\text{GroundedQA}} + \\
\alpha\underbrace{\text{argmax}_{q\in Q}\Psi(q^+|v_t, Q)\Phi (t | v, q^+)}_{\text{Grounding}},
\end{aligned}
\end{equation}
where $Q = Q^+\cup Q^-$ which comprises both the positive and negative questions of $v_t$ and $\alpha$ is a trade-off parameter. Note that the Grounding-term coarsely identifies the question-relevant video moment $t$, while the GroundedQA-term not only makes the prediction but also helps to refine the moment $t$ with answer supervision. The overall objective thus enforces the grounded video contents to be relevant to both the answers and the questions.

%% file: sections/experiment.tex
\section{Experiments}\label{sec:exp}
\subsection{Overview}
Our experiments answer three research questions: \textbf{Q1}: To what extent are the current VLMs' predictions %are truly 
grounded on \emph{relevant} video content? \textbf{Q2}: Does better QA performance imply
better grounding %results 
and vice versa? \textbf{Q3}: How effective is our Gaussian masking %grounding 
mechanism?
%As such, 
We study a wide variety of VLMs, covering different architectures (dual and stacked transformers), vision encoders (task-specific and pretrained with image- or video-text data), and text encoders (BERT, RoBERTa, DeBERTa, Flan-T5):
\begin{enumerate}
    \item \textbf{VGT} \cite{xiao2022vgt} is a task-specific, dual-style graph transformer model. 
    It encodes spatio-temporal object information \cite{ren2015faster} for VideoQA. We also investigate VGT with RoBERTa~\cite{liu2019roberta} as suggested by~\cite{xiao2023contrastive}.
   \item \textbf{Temp[Swin]} is a dual architecture. The Swin Transformer (SWT) \cite{liu2021swin} is pre-trained on ImageNet~\cite{deng2009imagenet}.
   \textbf{Temp[CLIP]} and \textbf{Temp[BLIP]} follow the same dual architecture, but use ViT \cite{dosovitskiyimage} pretrained by CLIP \cite{radford2021learning} and BLIP \cite{li2022blip} respectively as vision encoders. 
   \item \textbf{VIOLETv2} \cite{fu2022empirical} adopts a stacked transformer. It uses video Swin Transformer \cite{liu2022video} (VSWT) and BERT for vision and text encoding, respectively. The model is pretrained with both image- and video-text data, and achieves SoTA on various VL tasks.
   \item \textbf{FrozenBiLM} \cite{yang2022zero} applies a stacked transformer. It uses CLIP as vision encoder and highlights the strength of adapting \emph{frozen} large language models (LLMs) (\eg, DeBERTa-V2-XL (1B) \cite{hedeberta}) for VideoQA. 
   \item \textbf{IGV} \cite{li2022invariant} and \textbf{SeViLA} \cite{yu2023self} are additionally reproduced for comparison. Both works emphasize grounding keyframes for VideoQA. IGV is built on a visual graph, whereas SeViLA is founded on BLIP-2 \cite{li2023blip}. It exploits ViT-G \cite{zhai2022scaling} and \emph{frozen} LLM (\eg, Flan-T5-XL (3B)\cite{chung2022scaling}) for video localization and QA. In our implementation, we choose the smallest time spans that can enclose the localized keyframes as grounded moments.
   % It achieves SoTA QA performance across multiple benchmarks.
\end{enumerate}
\textbf{Experimental Settings}. For all the models, we uniformly sample 32 frames from each video and freeze the vision encoders. In \emph{post-hoc} analysis, the temporal attention thresholds are set dynamically according to %determined by searching around the 
mean attention values to maximize the grounded QA accuracy.
The number of negative questions in Eqn.~\ref{eqn:cgvqa} is kept the same as the number of distractor answers in MCQA to facilitate joint optimization. The trade-off parameter $\alpha$ is set to 1 and 0.1 for dual and stacked transformers, respectively. During inference, the hyperparameter $\gamma$ for the Gaussian confidence interval is chose from \{1, 0.8\} depending on different models. Our final results are reported based on a combination of predictions from Gaussian and temporal attention. All hyper-parameters are tuned on the validation set, and unless otherwise indicated, the results are reported on the test set. Other details are described in Appendix~\ref{sec:appimp}. 
% Other training details are 

\textbf{Evaluation}. We report accuracy for QA \cite{zhong2022Video}, which stands for the percentage of correctly answered questions. For visual evidence grounding, we use intersection over prediction (IoP) to measure whether the predicted temporal window lies inside the ground truth. Additionally, we include temporal IoU following video grounding benchmarks. 
For both IoP and IoU, we report the mean and the values with overlap thresholds of 0.3 and 0.5. If a QA pair involves multiple temporal segments, we report the results based on the one with maximal overlap with the prediction. Notably, we define grounded QA accuracy (Acc@GQA) to inspect the percentages of questions that are correctly answered and also visually grounded (\ie, IoP $\geq$ 0.5).

\subsection{Result and Analysis}

\subsubsection{\textbf{Q1}:Are the answers visually grounded?} 
We focus on Acc@QA, Acc@GQA and IoP@0.5 in the \emph{Post-hoc} (PH) block of Tab.~\ref{tab:main_res}. Generally, %the results reveal that 
the existing VLMs excel at QA but are weak %to visually substantiate their 
in grounding the answers in the videos. For example, all the methods exceed 50\% in QA accuracy, yet cannot reach more than 12-16\% for grounded QA accuracy. In fact, the SoTA QA model (FrozenBiLM) achieves an accuracy of 69\% for QA compared to a surprisingly low 16\% for GQA. The results of IoP@0.5 suggest that the large disparity is mainly due to the models' poor performance in temporal grounding.  It is also due partly to inconsistency between grounding and QA because not all correct grounding yields correct answers according to Acc@GQA \vus IoP@0.5.
We additionally exclude the influence of sparse video sampling by investigating the coverage of QA content w.r.t the number of sampled video frames in Fig.~\ref{fig:gdqa}(a). The figure shows that the sampled 32 frames can cover almost all QA contents. Moreover, to understand the extent of such poor performance,
we estimate the upper-bound performance via a human study on 10\% of the test data.  The study shows that the participants   
correctly answered 93\% of the questions, with 82\% being visually grounded. 

Given the above observations, we believe most of these models' answers are not grounded on the relevant video content.  Instead, they are more likely derived from {language shortcuts} or {spurious correlations} with irrelevant visual context. 
\setlength{\tabcolsep}{5.5pt}
\begin{table*}[t!]
  \caption{Grounded QA performance on NExT-GQA test set. $\dagger$: original NExT-QA. D/S: Dual/Stacked. CM: Cross-modal pretrain. BT: BERT. RBT: RoBERTa. DBT: DeBERTa-V2-XL. FT5: Flan-T5-XL. Random: always choose the same answer id and return the whole video duration as grounding result. *: pretrain on video-language grounding dataset.}
  \label{tab:main_res}
  \vspace{-0.3cm}
  \centering
  \small
  \scalebox{0.8}{
  % \begin{tabular}{llccccccccccccc}
    \begin{tabular}{llcccclllllllll}
    \toprule
    \multicolumn{2}{c}{Model} & D/S & CM & Vision & Text  & Acc@QA & Acc@QA$^{\dagger}$ & Acc@GQA & mIoP & IoP@0.3 & IoP@0.5 & mIoU & IoU@0.3 & IoU@0.5 \\
    \midrule
    \rowcolor[gray]{.8}
    & Human & - & -& -& -  & 93.3 & - & 82.1 & 72.1 & 91.7 & 86.2 & 61.2 & 86.9 & 70.3   \\
    \rowcolor[gray]{.8}
    & Random & - & -& -& -  & 20.0 & 20.0 & 1.7 & 21.1 & 20.6 & 8.7 & 21.1 & 20.6 & 8.7   \\
    \midrule
     &IGV & - & N & ResNet & BT  & 50.1 & 51.3 & 10.2 & 21.4 & 26.9 & 18.9 & 14.0 & 19.8 & 9.6  \\
    &\textit{SeViLA}* & \textit{S} & \textit{Y} & \textit{ViT-G} & \textit{FT5} & \textit{68.1} & \textit{71.5} & \textit{16.6} & \textit{\textbf{29.5}} & \textit{\textbf{34.7}} & \textit{22.9} & \textit{\textbf{21.7}} & \textit{\textbf{29.2}} & \textit{\textbf{13.8}} \\
    \midrule
    \multirow{9}*{PH}
    & VGT & D & N & RCNN & BT  & 50.9 & 53.8 &12.7 & 24.7 & 26.0 & 24.6 & 3.0 & 4.2 & 1.4  \\
    & VIOLETv2 & S & Y & VSWT & BT &  52.9 & 57.2 & 12.8 & 23.6 & 25.1 & 23.3 & 3.1 & 4.3 & 1.3  \\
    & VGT & D & N & RCNN & RBT &  55.7 & 57.7 & 14.4 & 25.3 & 26.4 & 25.3 & 3.0 & 3.6 & 1.7  \\
    & Temp[Swin] & D & N & SWT & RBT &  55.9 & 58.7 & 13.5 & 23.1 & 24.7 & 23.0 & 4.9 & 6.6 & 2.3  \\
    & Temp[CLIP] &D & Y& ViT-B & RBT & 57.9 & 60.7 & 14.7 & 24.1 & 26.2 & 24.1 & 6.1 & 8.3 & 3.7  \\
    & Temp[BLIP] &D & Y& ViT-B & RBT & 58.5 & 61.5 & 14.9 & 25.0 & 27.8 & 25.3 & 6.9 & 10.0 & 4.5  \\
    & \cellcolor{green!15}Temp[CLIP] &\cellcolor{green!15} D & \cellcolor{green!15}Y & \cellcolor{green!15}ViT-L & \cellcolor{green!15}RBT & \cellcolor{green!15}59.4 & \cellcolor{green!15}62.5 & \cellcolor{green!15}15.2 & \cellcolor{green!15}25.4 & \cellcolor{green!15}28.2 & \cellcolor{green!15}25.5 & \cellcolor{green!15}6.6 & \cellcolor{green!15}9.3 & \cellcolor{green!15}4.1  \\
    & \cellcolor{red!15}FrozenBiLM & \cellcolor{red!15}S & \cellcolor{red!15}Y & \cellcolor{red!15}ViT-L & \cellcolor{red!15}DBT &\cellcolor{red!15}69.1 & \cellcolor{red!15}71.8 & \cellcolor{red!15}15.8 & \cellcolor{red!15}22.7 & \cellcolor{red!15}25.8 & \cellcolor{red!15}22.1 & \cellcolor{red!15}7.1 & \cellcolor{red!15}10.0 & \cellcolor{red!15}4.4  \\
    \midrule
    \multirow{1}*{NG}
    & \cellcolor{green!15}Temp[CLIP] & \cellcolor{green!15}D& \cellcolor{green!15}Y& \cellcolor{green!15}ViT-L & \cellcolor{green!15}RBT & \cellcolor{green!15}59.4 & \cellcolor{green!15}62.7 & \cellcolor{green!15}15.5 & \cellcolor{green!15}{25.8} & \cellcolor{green!15}28.8 & \cellcolor{green!15}{\bf25.9} & \cellcolor{green!15}7.7 & \cellcolor{green!15}10.9 & \cellcolor{green!15}4.6  \\
    & \cellcolor{red!15}FrozenBiLM & \cellcolor{red!15}S & \cellcolor{red!15}Y & \cellcolor{red!15}ViT-L & \cellcolor{red!15}DBT & \cellcolor{red!15}70.4 & \cellcolor{red!15}73.1 & \cellcolor{red!15}17.2 & \cellcolor{red!15}24.0 & \cellcolor{red!15}28.5 & \cellcolor{red!15}23.5 & \cellcolor{red!15}9.2 & \cellcolor{red!15}13.0 & \cellcolor{red!15}5.8  \\
    \midrule
    \multirow{2}*{NG+}
    & \cellcolor{green!15}Temp[CLIP] &\cellcolor{green!15}D & \cellcolor{green!15}Y& \cellcolor{green!15}ViT-L & \cellcolor{green!15}RBT & \cellcolor{green!15}60.2\textcolor{red}{\footnotesize{+0.8}} & \cellcolor{green!15}63.3\textcolor{red}{\footnotesize{+0.8}} & \cellcolor{green!15}16.0\textcolor{red}{\footnotesize{+0.8}} & \cellcolor{green!15}25.7\textcolor{red}{\footnotesize{+0.3}} & \cellcolor{green!15}{31.4}\textcolor{red}{\footnotesize{+3.2}} & \cellcolor{green!15}25.5\textcolor{red}{\footnotesize{+0.0}} & \cellcolor{green!15}{12.1}\textcolor{red}{\footnotesize{+5.5}} & \cellcolor{green!15}{17.5}\textcolor{red}{\footnotesize{+8.2}} & \cellcolor{green!15}{8.9}\textcolor{red}{\footnotesize{+4.8}}  \\
    & \cellcolor{red!15}FrozenBiLM & \cellcolor{red!15}S & \cellcolor{red!15}Y & \cellcolor{red!15}ViT-L & \cellcolor{red!15}DBT &  \cellcolor{red!15}{\bf70.8}\textcolor{red}{\footnotesize{+1.7}} & \cellcolor{red!15}{\bf73.1}\textcolor{red}{\footnotesize{+1.4}} & \cellcolor{red!15}{\bf17.5}\textcolor{red}{\footnotesize{+1.7}} & \cellcolor{red!15}24.2\textcolor{red}{\footnotesize{+1.5}} & \cellcolor{red!15}28.5\textcolor{red}{\footnotesize{+2.7}} & \cellcolor{red!15}23.7\textcolor{red}{\footnotesize{+1.6}} & \cellcolor{red!15}9.6\textcolor{red}{\footnotesize{+2.5}} & \cellcolor{red!15}13.5\textcolor{red}{\footnotesize{+3.5}} & \cellcolor{red!15}6.1\textcolor{red}{\footnotesize{+1.7}}  \\
    \bottomrule
    \end{tabular}
  }
  % \vspace{-0.4cm}
\end{table*}
\textbf{To investigate the language shortcut}, we conduct a BlindQA experiment, in which we train only the language counterparts of the VQA models without video inputs. 
Tab.~\ref{tab:res-aba}(a) shows that BlindQA achieves 80\% of the performance of standard VQA (NormalQA), \ie, 50.3\% \vus 59.4\% for the dual models and 56.7\% \vus 69.1\% for the stacked models.
\textbf{To study the spurious correlations}, we test the VLMs by directly sampling inside (PosQA) or outside (NegQA) the ground-truth video segments. 
Surprisingly, the models' QA performances remain almost unaffected compared to a normal uniform sampling (NormalQA), likely because the image representations are not fine-grained enough to differentiate different frames. 
Tab.~\ref{tab:res-aba}(a) shows that providing the ground-truth temporal segments (PosQA) brings marginal improvement ($<$1\%) for the dual-style models and even hurts
stacked-style transformers, likely due to a distribution shift in visual inputs. Furthermore, excluding the temporal segments (NegQA) degenerates the performance by less than 1\% for both dual and stacked-style models. The above studies reinforce our belief that the current VLM's predictions are often not visually-grounded. 
% on relevant video contents, and instead relied on language shortcuts and a superficial vision-language correlation.
\setlength{\tabcolsep}{7pt}
\begin{table}
\caption{Performances under different settings. (+): with NG+. VQA: Question subset that BlindQA cannot answer. GDQA: Subset that both BlindQA and NegQA cannot answer but PosQA can.}
\vspace{-0.3cm}
\label{tab:res-aba}
\centering
\begin{subtable}{\linewidth}
  \centering
  \small
  \caption{}
  \scalebox{0.8}{
  \begin{tabular}{llcccc}
    \toprule
     \multicolumn{2}{c}{Model}  & NormalQA  & BlindQA & PosQA & NegQA \\
    \midrule 
    \multirow{2}*{Post-hoc} 
    & Temp[CLIP] & 59.4 & 50.3 & 59.8 & 59.1 \\
    &FrozenBiLM & 69.1 & 56.7 & 68.5 & 68.2  \\
    \midrule
    \multirow{2}*{NG+}
    &Temp[CLIP] & 60.2 & 50.3 & 61.0 & 59.4 \\
    &FrozenBiLM & 70.8 & 56.7 & 70.0 & 69.6  \\
    \bottomrule
  \end{tabular}
  }
\end{subtable}
% \vspace{10em}
\begin{subtable}{\linewidth}
  \centering
  \small
  \caption{}
  \scalebox{0.8}{
    \begin{tabular}{llllll}
    \toprule
     Models  & QA Set  & Acc@QA & Acc@GQA & mIoP & mIoU \\
    \midrule
    \multirow{5}*{\rotatebox[origin=c]{90}{Temp[CLIP]}}
    & Whole & 59.4 & 15.2 & 25.5 & 6.6 \\
    \cmidrule{2-6}
    & VQA & 35.7 & 9.7 & 25.2 & 7.0  \\
    & \cellcolor{green!15}VQA(+)& \cellcolor{green!15}39.4\textcolor{red}{\footnotesize{+3.7}} & \cellcolor{green!15}10.6\textcolor{red}{\footnotesize{+0.9}} & \cellcolor{green!15}25.5\textcolor{red}{\footnotesize{+0.3}} & \cellcolor{green!15}12.2\textcolor{red}{\footnotesize{+5.2}} \\
    & GDQA & 23.0 & 10.8 & 27.6 & 5.9     \\
    & \cellcolor{green!15}GDQA(+) & \cellcolor{green!15}30.2\textcolor{red}{\footnotesize{+7.2}} & \cellcolor{green!15}14.3\textcolor{red}{\footnotesize{+3.5}} & \cellcolor{green!15}29.3\textcolor{red}{\footnotesize{+1.7}} & \cellcolor{green!15}13.1\textcolor{red}{\footnotesize{+7.2}} \\
    \midrule
    \multirow{5}*{\rotatebox[origin=c]{90}{FrozenBiLM}}
    & Whole & 69.1 & 15.8 & 22.7 & 7.1 \\
    \cmidrule{2-6}
    & VQA & 47.6 & 11.2 & 22.2 & 6.6   \\
    & \cellcolor{red!15}VQA(+)&\cellcolor{red!15}50.0\textcolor{red}{\footnotesize{+2.4}} &\cellcolor{red!15}12.8\textcolor{red}{\footnotesize{+1.6}} &\cellcolor{red!15}23.7\textcolor{red}{\footnotesize{+1.5}} & \cellcolor{red!15}9.5\textcolor{red}{\footnotesize{+2.9}} \\
    & GDQA & 42.6 & 14.8 & 24.6 & 7.3 \\
    & \cellcolor{red!15}GDQA(+) &\cellcolor{red!15}44.0\textcolor{red}{\footnotesize{+1.4}} &\cellcolor{red!15}16.6\textcolor{red}{\footnotesize{+1.8}}&\cellcolor{red!15}27.0\textcolor{red}{\footnotesize{+2.4}} &\cellcolor{red!15}13.2\textcolor{red}{\footnotesize{+5.9}} \\
    \bottomrule
    \end{tabular}
    }
\end{subtable}
\vspace{-0.5cm}
\end{table}
\vspace{-0.4cm}
\subsubsection{\textbf{Q2}: Does better QA imply better grounding?}
\begin{figure*}[t!]
  \centering
    \includegraphics[width=1.0\linewidth, height=0.4\linewidth]{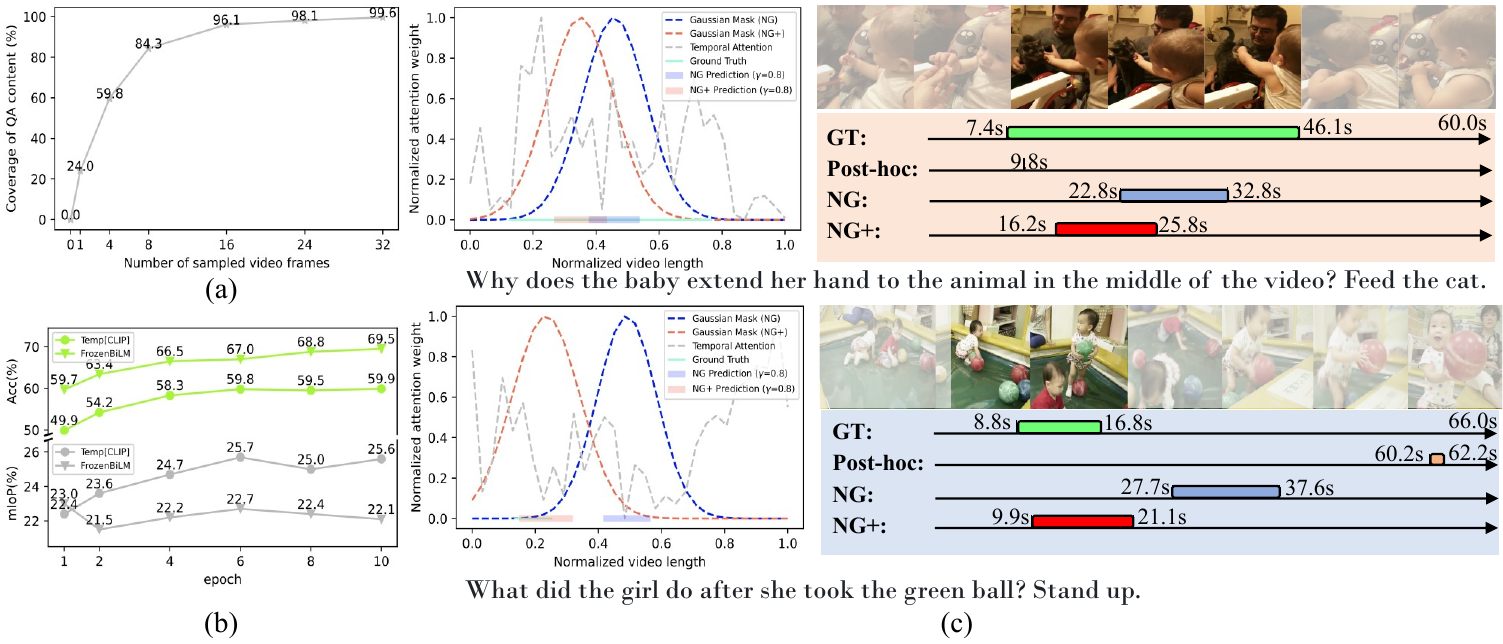}
  \caption{Analysis of visually-grounded VideoQA. (a) Coverage of QA content \wrt number of sampled video frames. (b) VQA and VG results \wrt training epochs on NExT-GQA Val set. (c) Visualization of the prediction examples. (Please zoom in for better view.)}
  \label{fig:gdqa}
  \vspace{-0.4cm}
\end{figure*}

First, by focusing on Acc@QA, mIoP and mIoU in Tab.~\ref{tab:main_res}. we find that \textbf{better QA is not necessarily established by better grounding, and the results vary across architectures}. 
For instance, by comparing across different architectures, FrozenBiLM shows the strongest QA performance, yet with surprisingly poor grounding, \eg, the IoP values are even worse than those of VGT which displays the lowest QA results among other transformer models.
This could be due to FrozenBiLM's freezing of the LLMs, causing its predictions to heavily rely on the \emph{common sense knowledge} of the LLMs rather than the provided videos (similar problem is also found on SeViLA). In contrast, VGT is a task specific model. It focuses on exploiting the fine-grained video information, and thus conditions better on the visual content. 
By comparing among different instantiations of the same architectures (\eg, Temp[Swin] to Temp[CLIP]) as well as different training epochs of the same models in Fig.~\ref{fig:gdqa}(b), we find that \textbf{the grounding performance (mIoP) improves along with the increase of QA accuracy for dual-style architectures yet not for stacked-style ones}.
Second, regarding the influence of grounding on QA, our conclusion is that \textbf{having grounding is better than not having it}. Yet, this is not controlled and opts for the underlying shortcuts %of the models convey and thus leaves for the 
when the models are allowed to learn freely. %to learn on their own. 
The conclusion is backed by the observations that PosQA always outperforms NegQA in Tab.~\ref{tab:res-aba}(a) regardless of model architectures. Moreover, our effort to improve grounding also brings better QA performance (Tab.~\ref{tab:main_res} $\&$ \ref{tab:res-aba}(b)). However, as mentioned, \textbf{correct grounding does not guarantee correct answers}.
\vspace{-0.4cm}
\subsubsection{\textbf{Q3}: Is Gaussian masking solution effective?}
We incorporate our Gaussian grounding mechanism (NG and NG+) into the top-performing dual- and stacked-style models and compare with
\emph{Post-hoc} baseline.
\footnote{Despite the weaker performance, we highlight the higher efficiency of dual-style implementation, especially in retrieval-based QA systems as exemplified by multi-choice QA.} 
Tab.~\ref{tab:main_res} shows 
that both \emph{NG} and \emph{NG+} lead to better grounding and QA performance. Also, NG+ generally outperforms NG, especially for dual-style architectures. Additionally, Tab.~\ref{tab:res-aba}(b) indicates that our superiority gets enlarged in answering the subset of questions that necessitate videos and temporal grounding. 

For better understanding, %To understand this enhancement, 
we analyze two cases in Fig.~\ref{fig:gdqa}(c). The top example shows that the Gaussian masks (\emph{NG} and \emph{NG+}) are more focused on the relevant video moment than temporal attention,
thus bringing better grounding, especially for IoU.
The bottom example highlights the strength of NG+.  In this case, there are multiple visual instances that correspond to the answer ``\texttt{girl stands up}''. The correct instance is the one after the ``\texttt{girl takes the \textcolor{green}{green} ball}'', though the instance after ``\texttt{take the \textcolor{red}{red} ball}'' is more salient.
Both the \emph{Post-hoc} and \emph{Naive} methods are distracted because they are learned via answer supervision alone. In contrast, \emph{NG+} finds the correct grounding since it also optimizes the cross-modal correspondence between questions and video segments. More detailed analyses are presented in Appendix~\ref{sec:app-exp}.  

\subsubsection{Method Comparison}
Compared with a random baseline, all methods effectively perform grounded QA (refer to Acc@GQA and IoP@0.5 in Tab.~\ref{tab:main_res}). More concretely, we find that both IGV and SeViLA obtain lower GQA accuracy than FrozenGQA though they also incorporate a sense of grounding in their models. The weakness manifest in both visual evidence grounding (IoP@0.5) and QA. However, we find that SeViLA performs much better than other methods in standalone grounding (mIoP and mIoU). We speculate this is because SeViLA is pretrained with localization supervisions \cite{lei2021detecting}. The observations thus point to possible future improvement by pretraining with location supervisions. Furthermore, they call for improved coordination between QA and grounding.
\vspace{-0.3cm}
\subsubsection{Other Observations}
Tab.~\ref{tab:main_res} also compares the Acc@QA  performance on NExT-GQA versus the full (original) NExT-QA test set. There is a consistent 2-3\% higher accuracy on the full set, suggesting that the questions 
rooted in local video moments are harder to answer than those %asking the 
rely on overall video content. Besides, 
the \textbf{cross-modal pretrained representations perform better than the uni-modal pretrained ones} for both VQA and visual grounding. Also, the image-text pretrained representations outperform those pretrained with video-text data.
Moreover, \textbf{existing dual-style architectures tend to have better grounding performance
than stacked ones} (Note that FrozenBiLM's high-ranking GQA result is due to its strong QA performance but not grounding). 
This is surprising, as there is no cross-modal interaction in dual-style implementations.   
We speculate that cross-modal transformers likely suffer from a 
\emph{uni-modal bias}, which leads to the attention being skewed towards the language side for predicting textual answers. 
The findings on the one hand consolidate the benefits of harnessing foundation VLMs or LLMs for videoQA. On the other hand, they accentuate the need to balance
between vision fact and text knowledge.

%% file: sections/conclusion.tex
\section{Conclusion}\label{sec: conclusion}
We summarize the following points and raise them as open challenges for the rest of the community:
\textbf{First}, current VLMs built on powerful language models excel in answering visual questions. Yet, their predictions often lack a strong connection to the pertinent visual information but instead heavily rely on languages short-cut and irrelevant visual context. This calls for more efforts towards the interpretability and trustability. \textbf{Second}, our experiments show that, localizing the questions, especially those featuring temporal actions and events is still a difficult and open challenge. Our studies indicate that solving this problem would largely benefit visually-grounded VideoQA. \textbf{Third}, although our solution improves grounding and QA, there is still a large gap compared with human performance. This leaves ample opportunity for follow-up works. 
\textbf{Last} but not least, we highlight the significance of NExT-GQA and hope it can contribute towards advancement in these areas.

\textbf{Limitations.} The NG+ method demands more memory and time to train (Appendix~\ref{sec:appeff}). Besides, our analyses are focused on multi-choice QA (Appendix \ref{sec:appmcqa}).

% \section*{Acknowledgements}
{\footnotesize 
\noindent \paragraph{Acknowledgements}
This research is supported by NUS NExT++ Research Center. 
The research is also supported by the National Research Foundation, Singapore under its NRF Fellowship for AI (NRF-NRFFAI1-2019-0001). Any opinions, findings and conclusions or recommendations expressed in this material are those of the author(s) and do not reflect the views of National Research Foundation, Singapore.}

%% file: sections/appendix.tex
\section{Appendix}\label{sec:app}
% This appendix provides additional introduction to 1) dataset construction, 2) implementation details for the three kinds of visually-grounded VideoQA mechanisms: \emph{Post-hoc}, \emph{Naive} and \emph{GQA}, and 3) more investigations and result analyses.

\subsection{Dataset Construction}
\label{sec:appdset}
% We use Elan \cite{elan65} for annotation (see Fig.~\ref{fig:anno-tool}). 
% During annotation, each video is decoded at its default frame rate for watching.
% \begin{figure*}[h]
%   \centering
% \includegraphics[width=1.0\linewidth]{figures/annotool.png}
%   \caption{Our annotation interface.}
%   \label{fig:anno-tool}
%   \vspace{-0.3cm}
% \end{figure*}

\begin{figure*}[t!]
  \centering
  \includegraphics[width=1.0\linewidth]{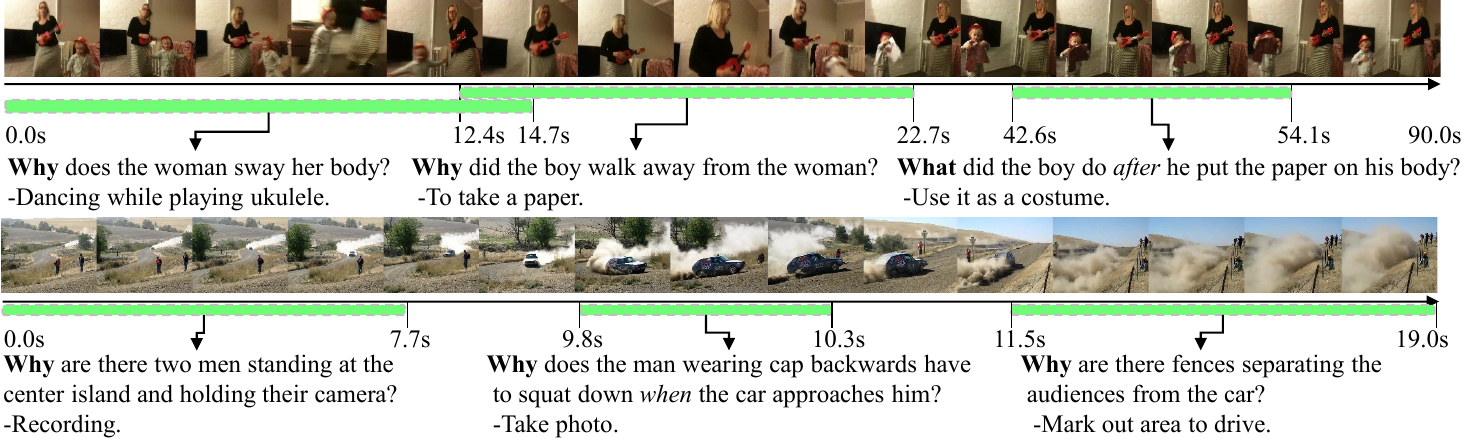}
  \caption{Examples of annotations in NExT-GQA. }
  \label{fig:seg-exp}
  % \vspace{-0.4cm}
\end{figure*}

\textbf{Criteria}.
We set clear criteria to limit the %resolve 
ambiguities and %to maximally reduce 
subjectiveness. 1) For each question, the %temporal 
annotation should %enclose all parts 
encompass the entire temporal segment %that can tell 
which features the answer and also %involve 
sufficient context to interpret the question. 2) If the visual content mentioned in the question %does not present at the same time or next to each other in time dimension
is not simultaneous or contiguous in time with the answer, then the annotation should focus on the answer. 3) If visual evidence for an answer appears multiple times in the video, then all relevant video moments (individual segment) should be annotated. 4) If the answer of the question can be seen throughout the entire video, the question is omitted. %then we omit this question. 
Yet, to ensure we can collect sufficient labels, we pay annotators on a per annotated segment basis.
Fig.~\ref{fig:seg-exp} shows two examples of our annotation outcome.
% according to the final number of annotated temporal labels. 

\subsection{Implementation Details}
\label{sec:appimp}
\textbf{Post-hoc.} 
For dual-style transformer, we have tried both \emph{attention-pooling} the visual tokens as well as \emph{prepending} a summary token and then averaging the multi-head attention of transformer. We find that the two methods bring similar QA performance. Yet, the \emph{prepending} approach demands much more training epochs and thus we chose attention-pooling as our final solution. Moreover,
to obtain a reasonable time span from the learned temporal attention, we treat the frame of maximal attention value as the pivot location, and search around it to enclose the frames whose attention values satisfying certain criteria. Before that, the attention values are normalized to [0, 1] using min-max method. The criteria of whether a frame should be enclosed are jointly determined by its attention score and its distance with the pivot frame. In our implementation, we also smooth the attention values and the distance threshold is set to 10s. Finally, the minimal frame id and the maximal frame id are mapped to the time seconds to obtain the temporal span. Note that the frame of maximal attention will always be selected. 

\textbf{Naive Gaussian (NG).} For both dual- and stacked-style architectures, the Gaussian prediction head is implemented with a lightweight transformer layer followed by linear projectors. Specifically, the Gaussian mask G (with dimension equal to the length of frames sequence $F$) is propagated to each self-attention head to weight the original self-attention weights before aggregating (summarizing) the value vectors, \ie, $F_h = G\cdot \text{softmax}(\frac{F^K(F^Q)^\top}{\sqrt{d_k}}) F^V$, in which Q, K, V indicate the respective query, key and value vector in self-attention. Notably, as there is no independent visual stream in stacked-style transformer, we pick the tokens belonging to the visual inputs and go through the Gaussian-weighted transformer. The resultant tokens are then preprended back into the multi-modal token sequence for answer prediction.

\begin{figure*}[t!]
  \centering
\includegraphics[width=1.0\linewidth]{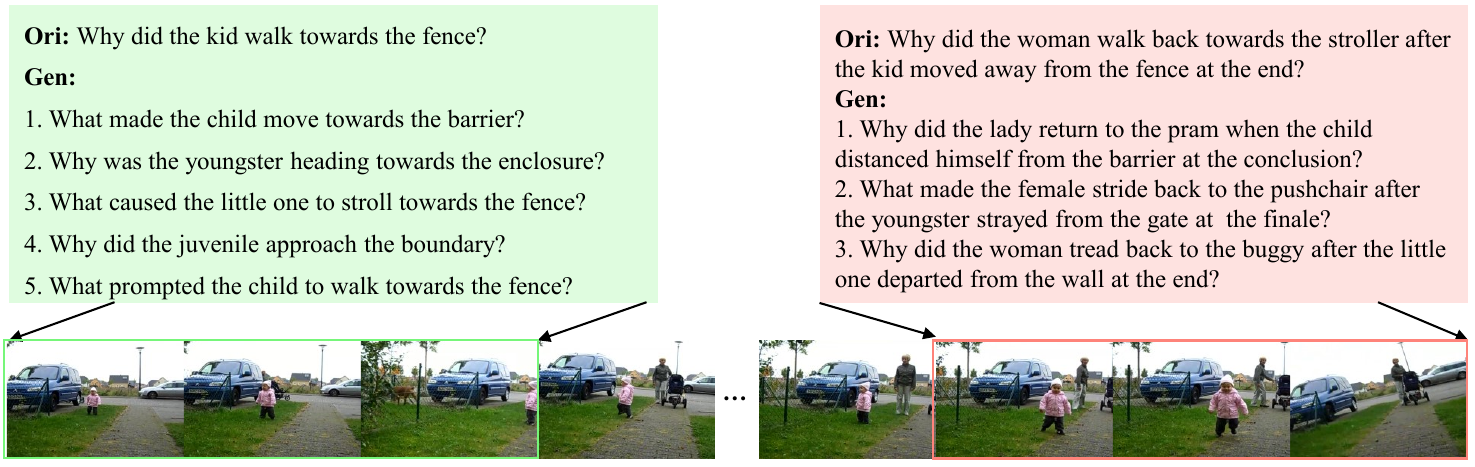}
  \caption{Examples of generated questions by GPT-4.}
  \label{fig:appgen}
  \vspace{-0.3cm}
\end{figure*}

\textbf{Video-Question Correspondence Learning (NG+).} 
We find that a two-stage training paradigm to pretrain with the Grounding-term and then finetune with both objectives in Eqn.~\ref{eqn:cgvqa} brings better performance than one-stage training. In both stage, the negative questions are selected from the same videos as the positive question at a chance of 0.3. Note that we exclude the descriptive questions because their answers usually appear throughout the video. Also at a chance of 0.3, we replace the positive question with a rephrased one. During generation, we prompt GPT-4 to focus on the nouns and actions in the questions, to ensure the generated questions reflect the same video moment with the original question. We show in Fig.~\ref{fig:appgen} some generated examples.

\textbf{Others.}
We train all models 10$\sim$20 epochs with initial learning rate of 1e-5. Earlier stopping is adopted if the validation results do not increase in 5 epochs. The batch size is set to 64 for dual-style models and 4$\sim$6 for stacked-style ones. During inference, to fuse the temporal windows derived from Gaussian masking and temporal attention, we simply choose the overlap area of two windows as the final prediction. If there is no overlap, we choose the predictions from temporal attention for better performance. 
% Unless otherwise specified, our major experiments are conducted on 4 A5000 GPUs with each 24G memories. 
% We find that FrozenBiLM with NG+ method takes more memory and thus conduct related experiment on 4 R8000 GPUs with each 48G memories. 
% The hyper-parameters are tuned on the validation set, and the final results are reported over the best results of 3 runs.

\subsection{Additional Experiments}
\label{sec:app-exp}
\subsubsection{Can multiple Gaussian masks help?}
We take Temp[CLIP] with Naive Gaussian (NG) grounding approach to study the effect of using different number of Gaussian masks. The results in Tab.~\ref{apptab:npro} show that using multiple Gaussian masks will hurt the QA accuracy though it increases the grounding performance according to IoU value. The best grounded QA (Acc@GQA) result is achieved by using 5 Gaussian masks. Nonetheless, the improvement over a single Gaussian mask is negligible, \eg, from 15.5\% to 15.8\%. Therefore, we by default use a single Gaussian mask in major experiments. This also brings higher efficiency. 
\setlength{\tabcolsep}{5.5pt}
\begin{table}[t!]
\centering
\small
\caption{Results of using different number of Gaussian masks.}
\label{apptab:npro}
\vspace{-0.3cm}
\scalebox{0.8}{
    \begin{tabular}{lccccc}
    \toprule
    Model & \#Masks & Acc@QA & Acc@GQA & mIoP & mIoU \cr
    \midrule
    \multirow{4}*{Temp[CLIP] (NG)}
    & 1 & 59.4 & 15.5 & 25.8 & 7.7 \cr
    & 3 & 57.9 & 15.2 & 25.6 & 9.1 \cr
    & 5 & 58.8 & 15.8 & 25.7 & 9.1 \cr
    & 7 & 58.2 & 15.4 & 25.7 & 10.9 \cr
    \bottomrule
    \end{tabular}
    }
    \vspace{-0.5cm}
\end{table}

\subsubsection{Does the generated questions help?}
We additionally study the effect of extended positive questions in the NG+ method. As shown in Tab.~\ref{apptab:gpt}, we find that it improves the QA results (Acc@QA) but not for grounded QA (Acc@GQA). In terms of grounding, it brings slightly higher IoU result yet lower IoP compared with the models without using the generated questions. We use the generated positive questions in our final experiments, considering that it improves QA and does not hurt grounded QA. The benefit could be more significant if we rephrase for more questions; currently, we only rephrase for 10\% of the questions in the training set. Yet, this will result in additional compute cost.
\setlength{\tabcolsep}{5.5pt}
\begin{table}[ht]
\centering
\small
\caption{Performances without (w/o) using generated questions.}
\label{apptab:gpt}
\scalebox{0.8}{
  \begin{tabular}{llcccc}
    \toprule
     \multicolumn{2}{c}{Model}  & Acc@QA  & Acc@GQA & mIoP & mIoU \\
    \midrule 
    \multirow{2}*{NG+} 
    & Temp[CLIP] (w/ Gen)& 60.2 & 16.0 & 25.7 & 12.1 \\
    & Temp[CLIP] (w/o Gen) & 59.3 & 16.0 & 26.7 & 9.9  \\
    \midrule
    \multirow{2}*{NG+}
    &FrozenBiLM (w/ Gen) & 70.8 & 17.5 & 24.2 & 9.6 \\
    &FrozenBiLM (w/o Gen) & 70.2 & 17.5 & 24.4  & 8.7  \\
    \bottomrule
  \end{tabular}
}
\vspace{-0.4cm}
\end{table}

\subsubsection{Model Efficiency}
\label{sec:appeff}
\setlength{\tabcolsep}{1.9pt}
\begin{table}[t]
\centering
\small
\caption{Model Efficiency.}
\label{tab:appeff}
\scalebox{0.8}{
  \begin{tabular}{lccccc}
    \toprule
    Model  & Train Param.  & Infer Param. & Model Size & Time (Train) & Time (Infer) \cr
    \midrule 
            Temp[CLIP] & 130.3M & 130.3M & 0.5G & 2.0m & 10.0s \cr
     \hspace{2em}w/ NG & 130.6M & 130.6M & 0.5G & 2.0m & 10.0s\cr
     \hspace{2em}w/ NG+ & 130.6M & 130.6M & 0.5G & 3.5m & 10.0s \cr
    \midrule
            FrozenBiLM & 29.7M & 1.2B & 3.8G & 0.3h & 1.0m \cr
    \hspace{2em}w/ NG  & 43.9M & 1.2B & 3.8G & 1.3h & 1.8m \cr
    \hspace{2em}w/ NG+ & 43.9M & 1.2B & 3.8G & 3.8h & 1.8m  \cr
    \bottomrule
  \end{tabular}
}
\vspace{-0.4cm}
\end{table}
We discuss the efficiency of Temp[CLIP] and FrozenBiLM in the visually-grounded QA task. For Temp[CLIP], all results are obtained with 1 A5000 GPU. For FrozenBiLM without NG+, the experiment was conducted on 4 A5000  GPUs; for FrozenBiLM with NG+, we run with 4 R8000 GPUs as the model needs about 46G per GPU memory. The time is reported based on 1 epoch over the training and validation data respectively. 
The results in Tab.~\ref{tab:appeff} show that our grounding module introduces little additional parameters for training and inference compared with the respective backbone models. Yet, the NG+ method takes more time to train. Another observation is that the Temp[CLIP] has much higher training and inference speed than FrozenBiLM. 
% This is because FrozenBiLM uses much larger language models.

\subsubsection{Generalization to Video-LLMs}
We study whether our grounding methods (post-hoc, NG and NG+) generalize to more recent multimodal large language models (MLLMs). We take Video-LLaMA \cite{zhang2023video} as an example. Video-LLaMA takes advantages of frozen LLaMA \cite{touvron2023llama} and pretrains Video Q-Former to bridge video inputs with LLaMA. It has demonstrated good VideoQA performance. To study its performance on NExT-GQA, we outline our adaptation as follows.

First, we omit the audio stream in Video-LLamA as NExT-GQA emphasizes visual grounding. Then, we find that the intermediate video Q-Former cuts off the direct correspondence between video frames/segments and answer outputs. This prevents a post-hoc analysis.
To circumvent the Q-Former yet also enjoy its cross-modal pretrained weights, 
we sample 32 video segments for each video and encode each segment by average-pooling the outputs of Q-Former. The segment representations, versus the original global Q-Former outputs, 
, are fed to LLaMA along with the QA texts (following the format in LLaMA-VQA \cite{ko2023large}) for answer decoding. Moreover, we summarize the Top-$K$ ($K=6$ is the maximal answer length) prediction scores of each video token as its confidence score for post-hoc temporal analysis. Besides, we prepend a special token to the video token sequence to predict the Gaussian parameters. For NG+, the large model size prevents joint training the two terms (Eqn.3 of the main paper) on our server. As a remedy, we apply a two-stage paradigm by first training for question \emph{grounding} and then fine-tuning for \emph{grounded QA}. Finally, to study the effect of multimodal \emph{video} pretraining, we include a model variant by substituting the Q-Former representations of the segments with CLIP features of their middle frames.

Tab.~\ref{tab:appres} highlights the following observations of Video-LLaMA' behavior on NExT-GQA: \textbf{1)} NG and NG+ give consistent improvements over a post-hoc method. \textbf{2)} Like our existing findings in the main paper, there is a large gap between QA and GQA accuracy.
\textbf{3)} Pretrained Video Q-Former improves over image-text pre-trained CLIP for QA but not
video grounding. Tab.~\ref{tab:appcomp} gives a comparison between Video-LLaMA and the two major backbones (TempCLIP and FrozenBiLM) in the main paper. We find that Video-LLaMA indeed shows higher Grounded QA (GQA) performance than non-LLM method Temp[CLIP]. However, like FrozenBiLM, the higher GQA accuracy of Video-LLaMA is resulted from its strong QA performance but not because of better grounding. In addition, we find that Video-LLaMA generally performs worse than FrozenBiLM in this task. We believe this is because Video-LLaMA solves QA by exploiting the LLMs to generate the answer word by word, while FrozenBiLM directly classifies each candidate as correct or incorrect answer which is more tailored-made for multi-choice QA. Similar findings can be found in the FrozenBiLM \cite{yang2022zero} paper which emphasizes the superiority of bi-directional pretrained LLMs to generatively trained ones for classification-based VideoQA. 

\setlength{\tabcolsep}{4pt}
\begin{table}[t!]
    \footnotesize
    \centering
    \caption{Results on NExT-GQA validation set. $\dagger$: Full validation set of NExT-QA. *: 2-stage training.}
    \label{tab:appres}
    \vspace{-0.8em}
    \scalebox{0.8}{
        \begin{tabular}{l|l|ccccc}
        \hline
         Backbone & Method & Acc@QA & Acc@QA$\dagger$ & Acc@GQA & mIoP & mIoU \\ 
        \hline
        \multirow{4}*{\makecell{Video-LLaMA(7B) \\(CLIP-VIT)}}
         & Post-hoc &  63.3 & 65.1 & 15.6 & 23.0 & 8.3 \\
        ~& NG & 64.3 & 67.2 & 16.5 & 24.9 & {\bf11.4} \\
        ~& *NG+ & {\bf66.7} & {\bf69.8} & {\bf17.2} & {\bf25.2}& 10.5 \\
        \cline{2-7}
        ~& Improves & +3.4 & +4.7 & +1.6 & +2.2 &+2.2 \\
        \hline
        \multirow{4}*{\makecell{Video-LLaMA(7B)\\(VQ-Former)}}
         & Post-hoc &  66.0 & 68.4 & 15.5 & 21.2 & 5.3 \\
        ~& NG & 66.9 & 69.4 & {\bf18.2} & {\bf25.1} & {\bf7.3} \\
        ~& *NG+ & {\bf68.5} & {\bf71.4} & 17.4 & 24.1 & 6.8 \\
        \cline{2-7}
        ~& Improves & +2.5 & +3.0 & +1.9 & +2.9 &+1.5 \\
        \hline
        \end{tabular}
    }
    % \vspace{-0.5cm}
\end{table}

\setlength{\tabcolsep}{4pt}
\begin{table}[t!]
    \footnotesize
    \centering
    \caption{Comparison on NExT-GQA test set}
    \label{tab:appcomp}
    \vspace{-0.8em}
    \scalebox{0.8}{
        \begin{tabular}{l|l|ccccc}
        \hline
         Method & Backbone & Acc@QA & Acc@QA$\dagger$ & Acc@GQA & mIoP & mIoU \\ 
        \hline
        \multirow{4}*{NG}
         & TempCLIP(130M) &  59.4 & 62.7 & 15.5 & {\bf25.8} &7.7 \\
        ~& Video-LLaMA(7B) & 65.1 & 68.3 & 16.6 & 24.9 & 7.7 \\
        ~& FrozenBiLM(1B) & {\bf70.4} & {\bf73.1} & {\bf17.2} & 24.0 & {\bf9.2} \\
        \cline{2-7}
        \hline
        \multirow{4}*{NG+}
         & TempCLIP(130M) &  60.2 & 63.3 & 16.0 & {\bf25.7} & {\bf12.1} \\
         ~& Video-LLaMA(7B) & 67.3 & 70.6 & 17.1 & 24.5 & 11.0 \\
        ~& FrozenBiLM(1B) & {\bf70.8} & {\bf73.1} & {\bf17.5} & 24.2 & 9.6 \\
        \hline
        \end{tabular}
    }
    \vspace{-0.5cm}
\end{table}

\subsubsection{Result Visualization}
\label{sec:appvis}
\begin{figure*}[t!]
  \centering
\includegraphics[width=0.9\linewidth, height=1.25\linewidth]{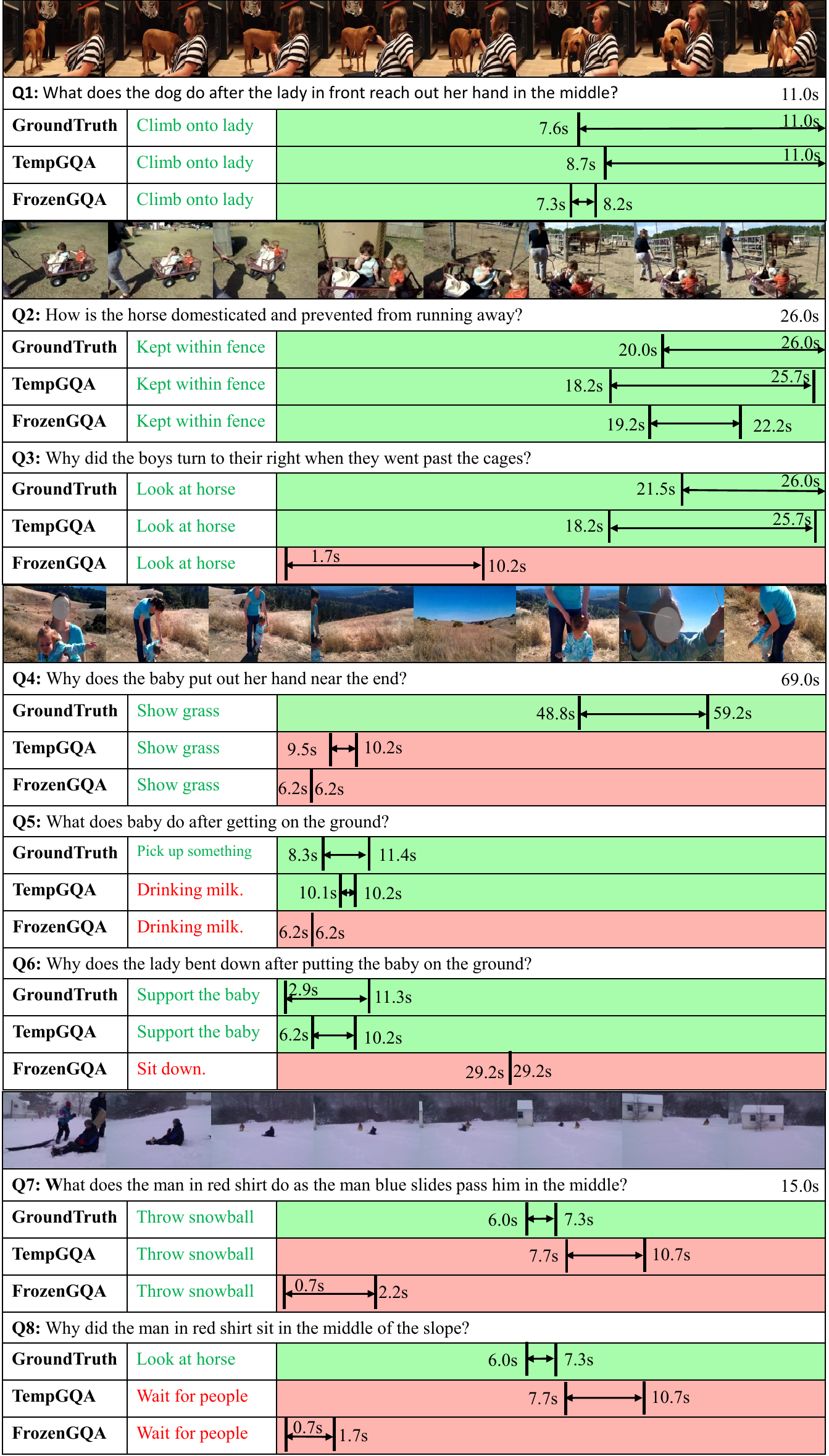}
  \caption{Result visualization on NExT-GQA. TempGQA and FrozenGQA denote Temp[CLIP] and FrozenBiLM with our NG+ grounding mechanism. The ground-truth and correct predictions are in green, while the wrong predictions are in red. }
  \label{fig:appvis}
\end{figure*}

We show some prediction cases in Fig.~\ref{fig:appvis}. Both models predict the correct answer with reasonable visual grounding results for Q1 and Q2. From the 3rd question, we show that the models suffer a lot in either correctly answering the questions (\eg, Q5, Q6 FrozenGQA and Q8) or providing the right visual evidence for the correct answers (\eg, Q3 FrozenGQA, Q4 and Q7). From the failure examples, we find that when the visual concepts in the answers present throughout the videos (\eg. ``\texttt{grass}" and ``\texttt{snow}" in Q4 and Q7 respectively), the models can easily predict the correct answers without the need to truly localizing the questioned video segments. Furthermore, the models are still weak in 1) answering the questions which involve small visual objects and 2) substantiating the answers when the visual evidence only takes small portion of the videos (Q4 $\sim$ Q8).

\subsection{Discussion on Multi-Choice QA}
\label{sec:appmcqa}
Popular open-ended VideoQA datasets, such as MSRVTT-QA, MSVD-QA and TGIF-QA, consist of very short videos, typically ranging from 3 to 15 seconds. They do not necessitate temporal grounding. While ActivityNet-QA contains long videos, a large portion of its questions are simple and can be answered with a single frame (by human). Given the above consideration, we experiment on NExT-QA, specifically on its multi-choice QA task as there is currently not much literature oriented for open-ended QA. Multi-choice QA tends to be more susceptible to language bias and spurious vision-language correlation. Because the provided negative answers may not always be distractive enough to challenge the selection of the correct answer without video consultation. Also, the visual concepts mentioned in the negative answers may not exist in the given videos at all. Conversely, our defined grounded-QA task would largely prevent or discourage such short-cut learning.